\documentclass[letterpaper]{article} 
\usepackage[utf8]{inputenc} 
\usepackage[T1]{fontenc} 

\usepackage{aaai2026}  
\usepackage{times}  
\usepackage{helvet}  
\usepackage{courier}  
\usepackage[hyphens]{url}  
\usepackage{graphicx} 
\usepackage{natbib}  
\usepackage{caption} 
\usepackage{algorithm}
\usepackage{algorithmic}

\usepackage{newfloat}
\usepackage{listings}
\DeclareCaptionStyle{ruled}{labelfont=normalfont,labelsep=colon,strut=off} 
\floatstyle{ruled}
\newfloat{listing}{tb}{lst}{}
\floatname{listing}{Listing}
\usepackage{multirow}
\usepackage{pifont}       
\usepackage{bbding}
\usepackage{algorithm}
\usepackage{algorithmic}
\usepackage{amsmath}
\usepackage{amssymb}

\nocopyright 

\title{MGFFD-VLM: Multi-Granularity Prompt Learning for Face Forgery Detection with VLM}
\author{
Tao Chen, Jingyi Zhang, Decheng Liu, Chunlei Peng
}

\usepackage{bibentry}

\begin{document}

\maketitle

\begin{abstract}
Recent studies have utilized visual large language models (VLMs) to answer not only "Is this face a forgery?" but also "Why is the face a forgery?" These studies introduced forgery-related attributes, such as forgery location and type, to construct deepfake VQA datasets and train VLMs, achieving high accuracy while providing human-understandable explanatory text descriptions. However, these methods still have limitations. For example, they do not fully leverage face quality-related attributes, which are often abnormal in forged faces, and they lack effective training strategies for forgery-aware VLMs.
In this paper, we extend the VQA dataset to create DD-VQA+, which features a richer set of attributes and a more diverse range of samples.
Furthermore, we introduce a novel forgery detection framework, MGFFD-VLM, which integrates an Attribute-Driven Hybrid LoRA Strategy to enhance the capabilities of Visual Large Language Models (VLMs).
Additionally, our framework incorporates Multi-Granularity Prompt Learning and a Forgery-Aware Training Strategy. By transforming classification and forgery segmentation results into prompts, our method not only improves forgery classification but also enhances interpretability. To further boost detection performance, we design multiple forgery-related auxiliary losses.
Experimental results demonstrate that our approach surpasses existing methods in both text-based forgery judgment and analysis, achieving superior accuracy.
\end{abstract}

\begin{figure}
  \centering
   \includegraphics[width=1\linewidth]{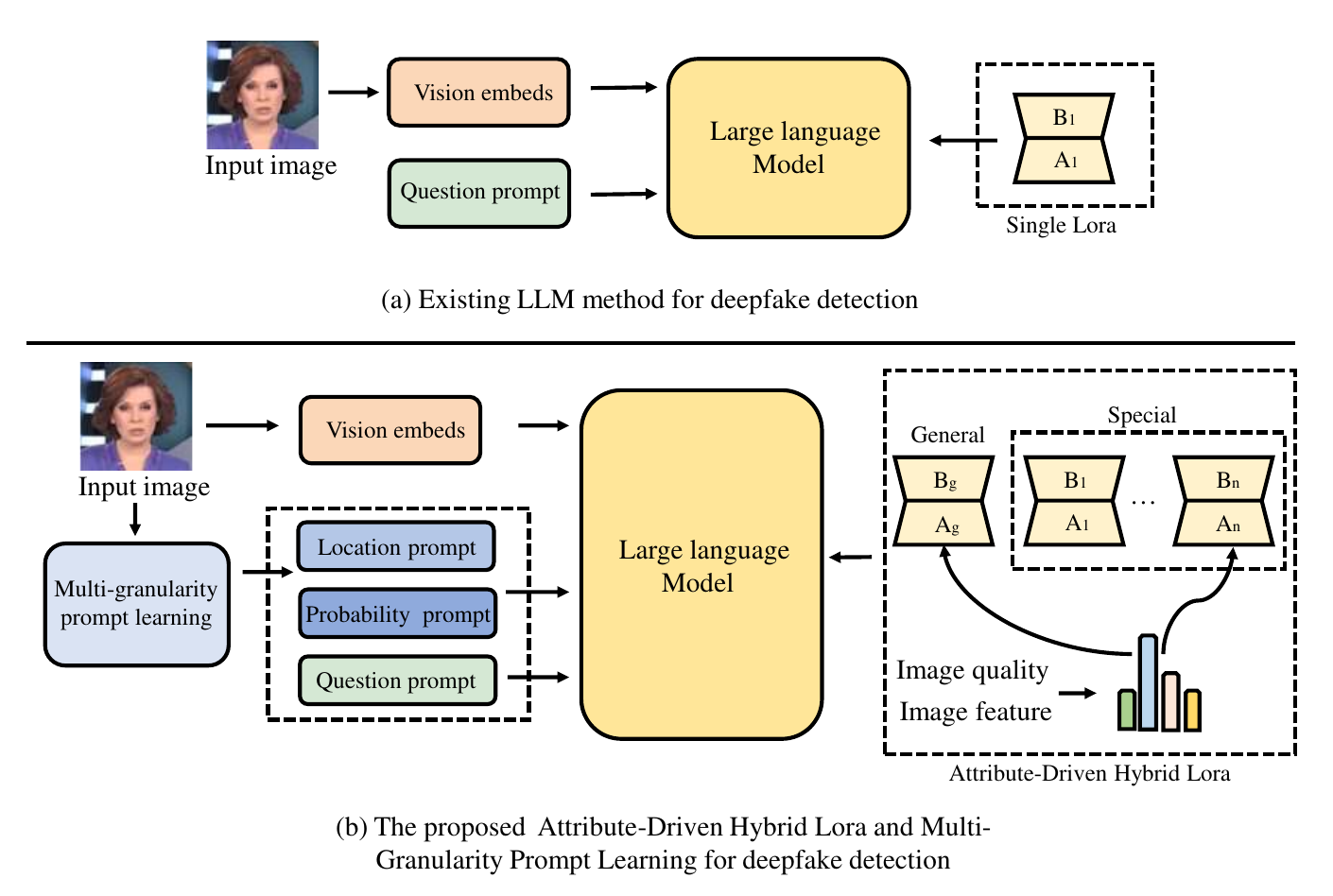}

   \caption{This figure illustrates the distinctions between our approach and the existing LLM method for deepfake detection. We employ attribute-driven  hybrid LoRA strategy and multi-granularity prompt learning to enhance the accuracy and interpretability of forgery explanations.}
   \label{fig:introduction}
\end{figure}

\section{Introduction}

Face forgery detection refers to distinguishing real faces from forgery faces through deep learning methods. With the increasingly mature development of AGI nowadays, most people cannot distinguish between real and forgery faces. Therefore, face forgery detection is currently a research hotspot and plays an important role in fields such as social security and economic security.

Existing face forgery detection models typically employ binary classification networks and leverage techniques like data augmentation, frequency analysis, and temporal analysis to improve performance. Although these models demonstrate excellent accuracy on classification metrics, they lack interpretability and fail to explain the basis of authenticity judgments.
Although saliency maps are a commonly used interpretability method, they can only show the decision - making regions and cannot explain why these regions are selected.
Beyond simply providing numerical predictions, understanding \textit{why} a face is identified as fake is crucial for developing robust face forgery detection systems.

\begin{figure}
  \centering
\includegraphics[width=1\linewidth] {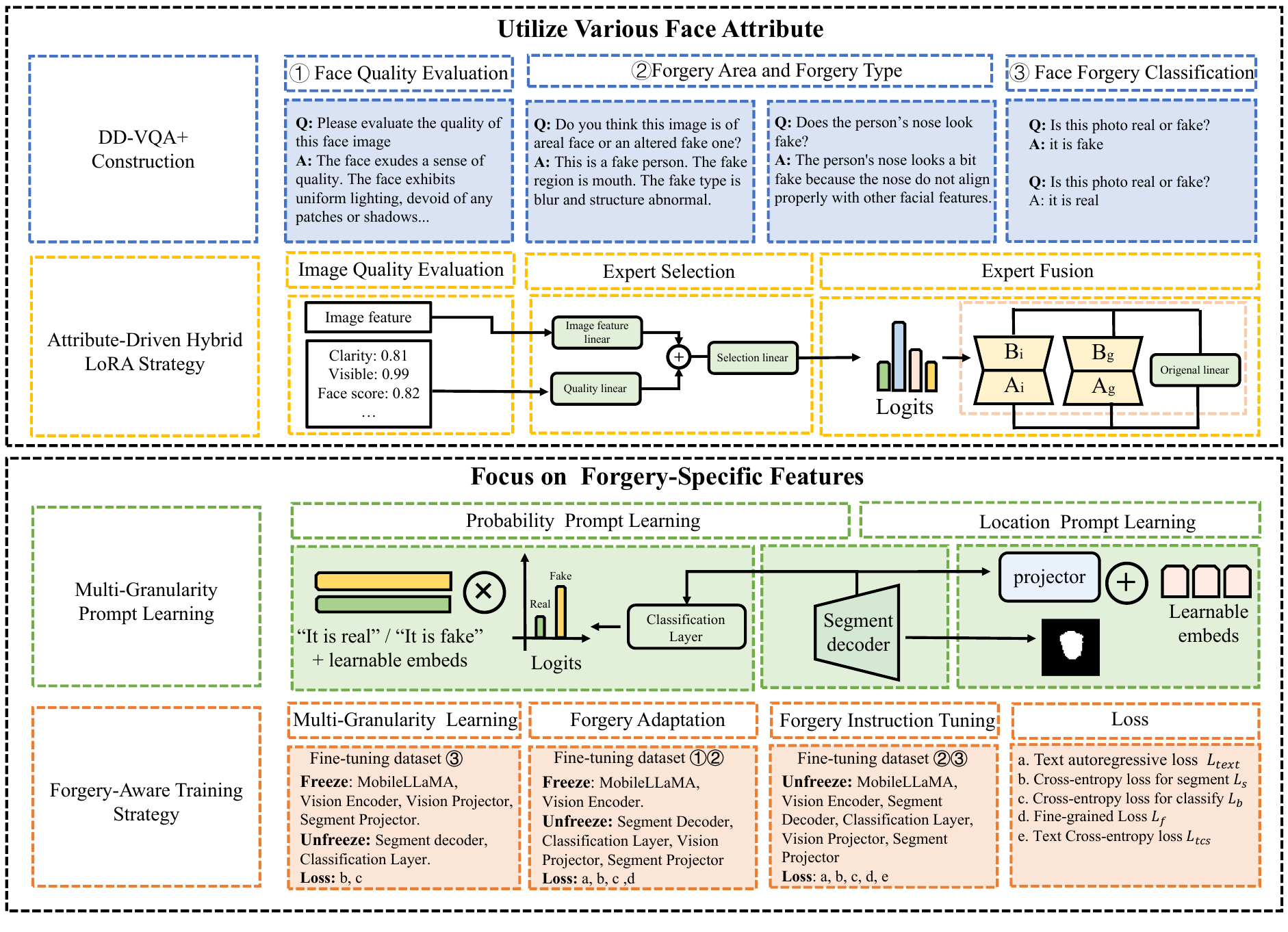}
   \caption{The overview of proposed MGFFD-VLM framework.}
   \label{fig:overview}
\end{figure}


To access this need, Zhang \textit{et al.}~\cite{zhang2024common} introduced a forgery VQA dataset(DD-VQA) based on human face perception and trained a question-answering visual large language model on this dataset to provide explanations for forgery detection. This approach has significantly advanced progress in the field.
However, these approaches still have two limitations: (1) Existing deepfake VQA datasets, despite their benefits, rely on manually annotated data. This limits their diversity and fails to account for other deepfake-related facial features, such as abnormal quality, clarity, visibility, and illumination, which often appear abnormal in forged faces.
(2) Given that forgery clues are often imperceptible, it is highly challenging to effectively guide large-scale visual-language models to focus on forgery-specific features.

\textbf{To address the issue (1),} we augment the text pair generation methods for Forgery Area and Forgery Type, which can be generated in a self-supervised manner based on DD-VQA. Meanwhile, we introduce additional facial features related to forgery detection to enhance diversity. Additionally, during the training process of large language models, we propose an Attribute-driven Hybrid LoRA Strategy to boost the performance of Visual Large Language Models (VLMs).
As shown in Figure~\ref{fig:introduction}, each sample selects an appropriate LoRA expert based on the current image quality and text characteristics, thus improving the detection capabilities in the varying image qualities.
\textbf{To address the issue (2),} we employ multi-granularity prompts to guide the model in capturing forgery-related details at various levels. As illustrated in Figure~\ref{fig:introduction}, this approach converts classification and forgery segmentation results into prompts. To further enhance the model's capability, in the Forgery-Aware Training Strategy, we introduce fine-grained image contrast losses and apply a soft-label cross-entropy loss at specific word prediction locations, thereby increasing the semantic distance between real and fake representations
The primary contributions of our work are as follows:


\begin{itemize} \item This paper introduces DD-VQA+, an enhanced version of DD-VQA, integrating forgery classification, localization, forgery-related attributes, and human-perceptible explanations. We supplement Forgery Area and Forgery Type, which can self-supervise the generation of forgery locations and attributes. Meanwhile, we introduce additional face quality-related attributes.
\item We propose Multi-Granularity Prompt Learning, enabling the VLM to leverage classification and forgery segmentation outputs as prompts, thereby enhancing both detection accuracy and model interpretability.
\item We propose an Attribute-driven Hybrid LoRA Strategy. Specifically, we dynamically select suitable LoRA experts based on image quality attributes and image features, thereby effectively analyzing forged images with varying attributes.
\item We propose Forgery-Aware Training Strategy. Increase the perception ability of forgery cues through a progressive learning strategy and carefully designed loss functions.
\end{itemize}

\section{Methodology}
Figure~\ref{fig:overview} provides an overview of the proposed \textbf{MGFFD-VLM} framework. Our approach has two major components: (i) better use of facial image attributes through an extended dataset and an attribute-driven LoRA design, and (ii) enhanced forgery-aware learning through multi-granularity prompts and a specialized training strategy. In the following, we first briefly introduce our base model (MobileVLM), then describe the construction of DD-VQA+, and finally detail our prompt learning, hybrid LoRA, and training strategy.



\begin{figure}
  \centering

   \includegraphics[width=0.8\linewidth]{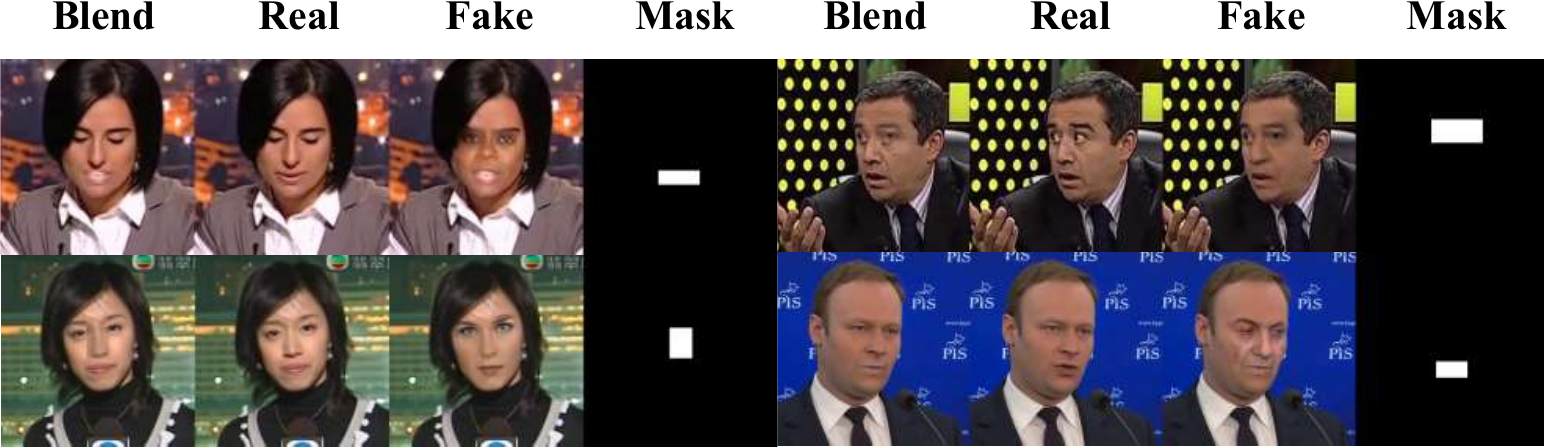}

   \caption{Demonstration of blend image generation in DD-VQA+ Construction. Random parts of real photos are replaced by fake faces.}
   \label{fig:local_vqa}
\end{figure}

\subsection{Mobile-VLM}
\begin{figure}
  \centering

   \includegraphics[width=1\linewidth]{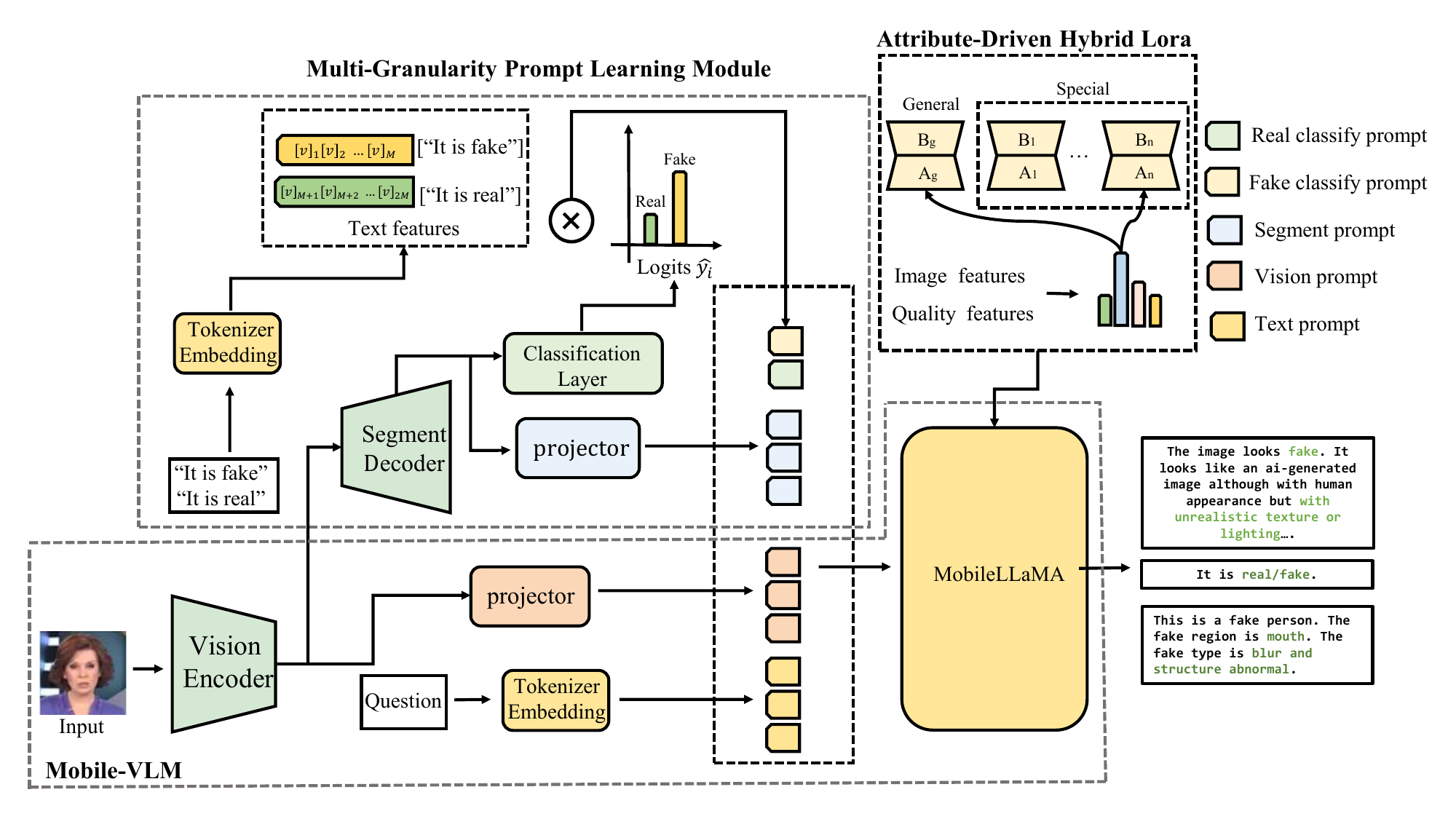}

   \caption{The overview of proposed Multi-Granularity Prompt Learning.}
   \label{fig:model architecture}
\end{figure}

We use Mobile-VLM, proposed by the Meituan team, as our baseline model. As shown in the lower part of Figure~\ref{fig:model architecture}, this model is primarily composed of three components: the Vision Encoder based on ViT-L/14, the Lightweight Downsample Projector, and the 1.4B parameter MobileLLaMA large language model. The Lightweight Downsample Projector, introduced by the Meituan team, significantly reduces the computational load, making it an efficient projector. Despite having only 1.7B parameters in total, this model achieves performance comparable to that of much larger models. Meanwhile, their training strategy consists of two stages. In the first stage, only the projector is unlocked, while in the second stage, fine-tuning of the large language model is carried out simultaneously.
For MobileVLM, it has inputs $(A, P, X)$, where $A$ is the expected answer, $P$ is the asked question, and $X$ is the corresponding image.
The model mainly extracts image features through ViT and maps the image features to the text space through a projector. Then, the question prompt $P^{question}$and the visual embeds $P^{vision}$ are input into the LLM together for prediction.

\subsection{DD-VQA+ Construction}
Zhang \textit{et al.}~\cite{zhang2024common} introduced a visual question answering (VQA) dataset for forgery detection, known as DD-VQA, which incorporates both common forgery questions \( Q_{common} \) and  annotated answers \( A_{common} \). The question \( Q_{common} \) prompts the model to evaluate whether specific facial features (such as the entire face, eyes, nose and mouth) appear fake in the image. All responses, \( A_{common} \), are manually annotated.
Their work demonstrated that manually annotated forgery locations and attributes can significantly improve the effectiveness of forgery detection, marking a major advancement in interpretable face forgery detection.
However, due to the limited scale of DD-VQA, we supplemented Forgery Area and Forgery Type, which can generate forgery locations and attributes in a self-supervised method.
Meanwhile, we argue that incorporating additional face quality-related attributes can further enhance VLMs' confidence in forgery judgment. We introduce more additional facial attributes related to face forgery detection in the Face Quality Evaluation. We extend DD-VQA to DD-VQA+ by incorporating the key facial attributes mentioned above. Specifically, we introduce three new attributes, which are described in detail below:

\textbf{Forgery Regions and Forgery Types.}
Accurately locating the forged areas can help large models better understand the concept of forgery.
we supplement Forgery Area and Forgery Type, which can generate forgery locations and attributes in a self-supervised.
Following the approach of Sun \textit{et al.}~\cite{sun2023towards}, we create blended images by replacing parts of real images with altered, fake segments, as illustrated in Figure~\ref{fig:local_vqa}. For a forgery image \( x_i \), we use the following formula:

\begin{equation}
\begin{aligned}
    x_{i}^{blend} = M_i \times x_{i}^{real} + (1 - M_i) \times x_{i}.
\end{aligned}
\end{equation}

$M_i$ is a binary mask that includes one of the following: the mouth, the nose, the eyes, or the entire face
, and $x_{i}^{real}$ represents the original image corresponding to $x_i$.
Unlike the original text, we randomly select regions instead of choosing the regions with the largest differences.
The standardized query $Q_{local}$ is defined as \textit{``Do you think this image is of a real face or an altered fake one?''}
The model's answer begins with the fixed statement $T_{label}$, \textit{``This is an example of a fake face''}, followed by details on the specific forgery area and type. For real faces, the model simply responds with \textit{``It is a real face.''} The response, $A_{local}$, can be composed of the following elements:

\begin{equation}
\begin{aligned}
    A_{local} = T_{label} + T_{region} + T_{type}.
\end{aligned}
\end{equation}

We predefine distinct text descriptions \( T_{region} \) for each replaced part.
There are five types of \(T_{type}\) in total, namely blur,structure abnormal, color difference, and blend boundary (details can be found in~\cite{sun2023towards}).

\textbf{Face Quality Evaluation.}
Since generative algorithms often neglect the consistency of illumination intensity and facial visibility, forged faces may exhibit abnormalities in these attributes.
To address this, we propose quantifying face quality indicators. The standardized prompt, $Q_{quality}$, is defined as \textit{``Please evaluate the quality of this face image.''} The model is then tasked with assessing the visual quality of the image from several perspectives, including overall impression, facial integrity, illumination intensity, illumination uniformity, clarity, and visibility. The response, $A_{quality}$, can be composed of the following components:

\begin{equation}
\begin{aligned}
    A_{quality} = T_{overall} + T_{face} + T_{intensity} \\
     + T_{uniformity}
    + T_{clarity} + T_{visibility}.
\end{aligned}
\end{equation}

In this formula, each \( T_{[\cdot]} \) represents a randomly chosen textual description corresponding to a specific aspect.
Taking \( T_{visibility} \) as an example, we first use the CPBD~\cite{narvekar2011no} metric to calculate the scores that indicate the face's visibility. These scores are then categorized into three levels, with corresponding prompts such as ``The facial visibility is high/mid/low.'' To ensure prompt diversity, we use GPT-4 to generate $50$ synonymous sentences with similar meanings. Finally, one sentence is randomly selected based on the score level. For example, when the CPBD score is 0.9, the generated prompt \( T_{visibility} \) will be ``The face is clearly visible.''
For the illumination factor \( T_{intensity} \), we follow a similar approach to the method in~\cite{ou2024clib} by training a CNN on the WIDER FACE dataset~\cite{yang2016wider} to determine the illumination score. Like \( T_{visibility} \), the score is divided into three levels, and GPT-4 is used to expand the text prompts. If \( T_{intensity} \) is less than 0.3, the prompt will be set to ``The brightness level on the face is dim.''
Further details can be found in the supplementary material.

\textbf{Face Forgery Classification.}
Additionally, we develop Face Forgery Classification attributes. Here, forgery judgment texts are generated in a self-supervised manner based on classification labels. The standardized prompt is ``Is this image real or fake?'', with corresponding answers provided for real and fake images, respectively.

For $x_i$, we can obtain:
\begin{equation}
\begin{cases}Q_i = \left\{ Q_{local}^i, Q_{common}^i, Q_{classify}^i, Q_{quality}^i \right\} \\A_i = \left\{A_{local}^i, A_{common}^i, A_{classify}^i, A_{quality}^i  \right\} \end{cases}.
\end{equation}

Here, $Q_i$ and $A_i$ are generated by randomly selecting elements from predefined lists according to the above formulas, enabling us to capture a broad range of characteristics and responses related to the input $x_i$.

\subsection{Multi-Granularity Prompt Learning}
Existing visual large language models (VLMs) lack visual prompts that are directly applicable to face forgery detection. To make the model focus more on forgery-related features, we introduce Multi-Granularity Prompt Learning to enhance both the accuracy and interpretability of forgery detection.
As shown in Figures~\ref{fig:model architecture}, our proposed prompt learning framework consists of two key components: the Probability Prompt and the Location Prompt. The Probability Prompt is generated by integrating binary classification logits with the corresponding text embeddings, while the Location Prompt is mapped to the text feature space via a projector, allowing it to capture spatial forgery cues more effectively.

\textbf{Probability Prompt Learning.}
We begin by extracting image features using the vision encoder, which are then passed to the segmentation decoder. We use the simple DeepLabv3 Decoder~\cite{yurtkulu2019semantic} as our segment Decoder. The segmentation features from the last layer are fed into a linear head to produce the binary classification result, denoted as $\hat{y_i}$.
Following the approach of CoOp~\cite{zhou2022learning}, we define the corresponding text embeddings as follows:
\begin{equation}
\begin{aligned}
\left\{
\begin{aligned}
P_{fake} &= [V]_1[V]_2 ... [V]_M [\text{it is fake}]\\
P_{real} &= [V]_{M+1}[V]_{M+2}... [V]_{2M} [\text{it is real}]
\end{aligned}
\right.
\end{aligned}
\end{equation}

Each $[V]_m$ ($m \in [1, 2,..., 2M]$) is a vector matching the word embedding dimension. $[\text{it is real}]$ and $[\text{it is fake}]$ are the associated text embeddings. We calculate the Probability Prompt as follows:
{\small
\begin{equation}
P_{i}^{probability} = F_{concat}([\hat{y_i} \times P_{fake} , (1-\hat{y_i}) \times P_{real}]).
\end{equation}
\small}

The higher the value of $\hat{y_i}$, the easier it is for the large language model to understand the prompt [it is fake]. Conversely, the lower the value of $\hat{y_i}$, the easier it is for the large language model to understand the prompt [it is real].

\textbf{Location Prompt Learning.}
To capture spatial forgery details, we obtain the segmentation feature from the last layer of the segmentation decoder. Then, through a projector, it is mapped to the feature space of the text, and we refer to this feature as $t^s_i$.
where the shape of $t_i^{s}$ is $L\times D$. Finally, $t_i^{s}$ is combined with $L$ learnable vectors for enhanced expressiveness:
\begin{equation}
P_{i}^{segment} = t_i^{s} + [V]_{2M+1}[V]_{2M+2}...[V]_{2M+L}. \end{equation}

Here, $M$, $H$ are hyperparameters, and $D$ is the feature dimension. In addition, the LLM input includes visual embeddings $P_{i}^{vision}$ aligned by projector.
Thus, the final LLM input $P_{i}^{all}$ incorporates the visual embeddings, question prompt, and both the probability and location prompts:
{\small
\begin{equation}
P_{i}^{all} = \left\{ P_{i}^{probability},P_{i}^{segment},P_{i}^{question},P_{i}^{vision}  \right
\}.
\end{equation}
\small}

\begin{figure}
  \centering
   \includegraphics[width=1\linewidth]{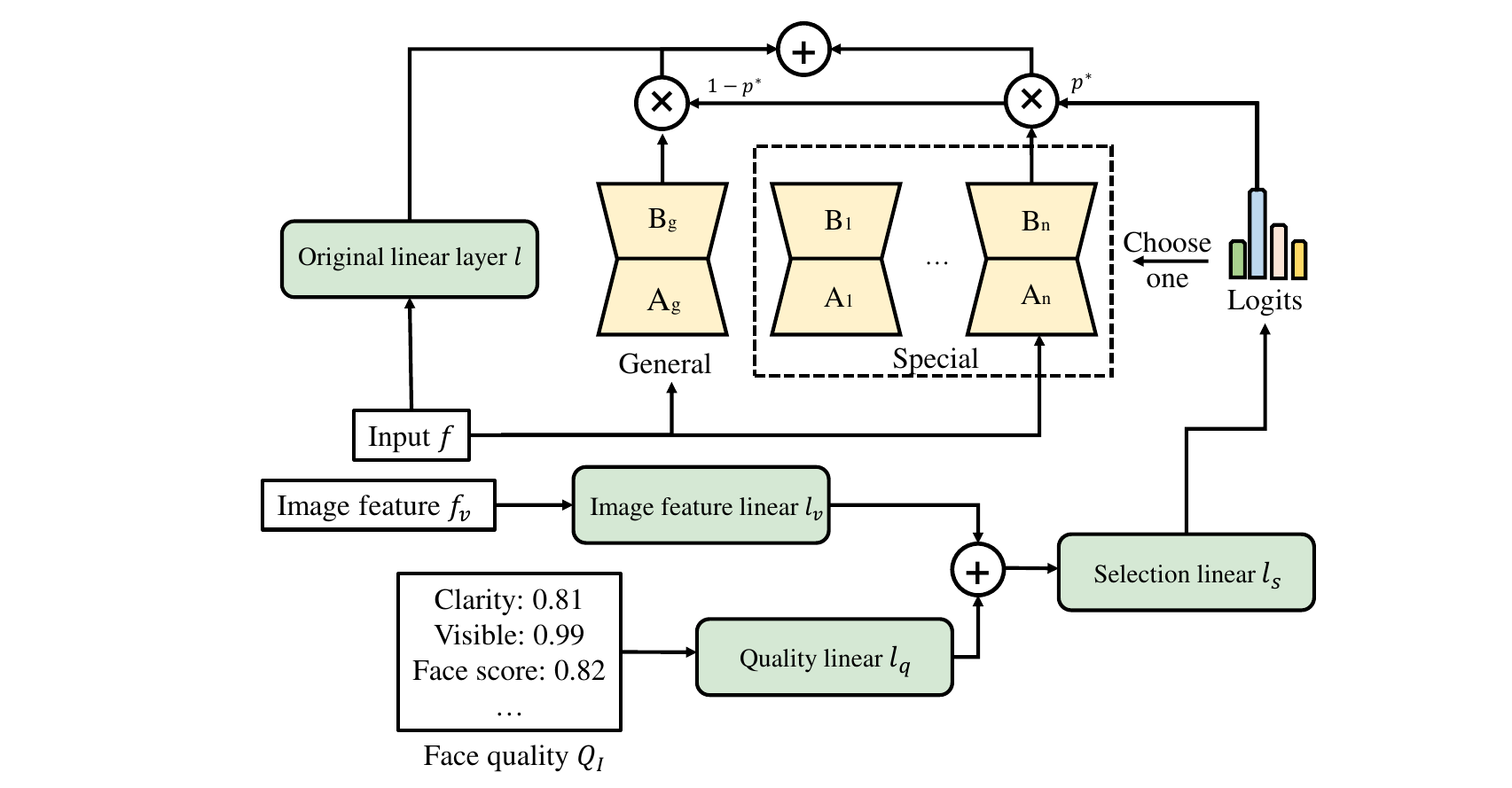}
   \caption{Illustration of the Attribute-Driven Hybrid LoRA Strategy.}
   \label{fig:lora}
\end{figure}

\subsection{Attribute-Driven Hybrid LoRA Strategy}
While face quality doesn't directly indicate forgery, it influences how artifacts appear. Our LoRA uses specialized experts to focus on different signals (e.g., high-frequency details or compression artifacts).
Inspired by~\cite{wu2023mole}, we propose an Attribute-Driven Hybrid LoRA Strategy that allows large models to focus on face quality scores while performing tampering-related question-answering tasks.
As illustrated in Fig.~\ref{fig:lora}, the process begins by employing a pre-trained image quality model \(M_q\) to extract a set of numerical quality indicators \(Q_I\) for each input image \(x\). These indicators are analogous to those used in standard Face Quality Evaluation. 


For each original linear layer \(l\) within the large language model, let the input feature from the previous layer be denoted as \(f\). This feature $f_v$ from vision encoder is processed by a image feature linear layer \(l_v\), which produces a feature-based selection representation \(v_s\). In parallel, the quality indicators \(Q_I\) is transformed into a quality-based selection representation \(q_s\) through a quality linear layer \(l_q\). These two representations are then concatenated and passed through a selection linear layer \(l_s\), resulting in a 4-dimensional vector. After applying softmax normalization, the resulting values represent the selection probabilities over the expert branches. The expert corresponding to the highest probability is chosen to process the input feature \(f\).

The final output of the hybrid LoRA module is computed as a weighted combination of the selected specific expert \(e_*\) (from a set of four specialized LoRA branches) and a global expert \(e_g\). Specifically, the updated output feature \(f'\) for the current layer \(l\) is calculated as:

\[
f' = p^* \cdot A_* B_*(f) + (1 - p^*) \cdot A_g B_g(f) + l(f)
\]

where \(p^*\) is the highest expert selection probability, \(A_*, B_*\) are the adaptation matrices of the selected specific expert, and \(A_g, B_g\) are those of the global expert. The original linear transformation \(l(f)\) is also incorporated via residual connection. The resulting output \(f'\) is then propagated to the next layer. Further implementation details are provided in the Supplementary Materials.

\begin{figure}
  \centering
   \includegraphics[width=0.6\linewidth]{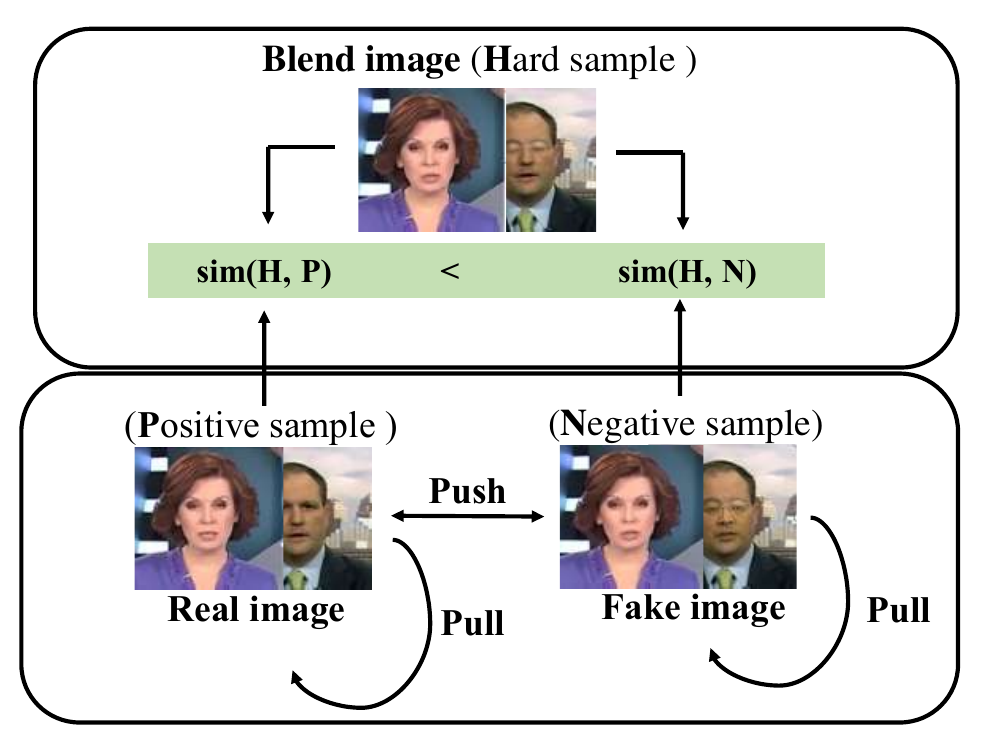}
   \caption{Illustration of the Fine-grained loss.  $P$ represents Real image, $N$ represents Fake image, and $H$ represents Blend image.  We use $sim(\cdot)$ to represent the cosine similarity between samples.}
   \label{fig:model-loss}
\end{figure}

\begin{table*}
\centering
\caption{We present experimental results from fine-tuning BLIP and BLIP-TI with the DD-VQA dataset, and Mobile-VLM and MGFFD-VLM with the extended DD-VQA+ dataset. Text accuracy and answer generation performance are reported for all four models. As our analysis relies on identifying the terms "real" or "fake" within the text, AUC scores are not provided. In this comparison, BLIP and BLIP-TI represent current state-of-the-art methods, Mobile-VLM serves as the baseline for DD-VQA+, and MGFFD-VLM is our proposed approach. The best results are shown in bold numbers.}

\label{table:compare1}
\scalebox{0.8}{
\begin{tabular}{ccccccccccc}
\hline
\multirow{2}{*}{Method} & \multirow{2}{*}{Dataset} & \multicolumn{4}{c}{Deepfake
  Detection} & \multicolumn{5}{c}{Answer
  Generation}           \\
\cline{3-11}
                        &                          & Acc$\uparrow$    & Recall$\uparrow$ & Precision$\uparrow$ & F1$\uparrow$      & BLUE-4$\uparrow$ & CIDEr$\uparrow$ & ROUGE\_L$\uparrow$ & METEOR$\uparrow$ & SPICE$\uparrow$  \\
\hline
BLIP~\cite{li2022blip}                    & DD-VQA                    & 0.8168 & 0.9596   & 0.7861     & 0.8642  & 0.3569  & 1.8177  & 0.5664    & 0.3301  & 0.6658  \\
BLIP-TI~\cite{zhang2024common}                 & DD-VQA                    & 0.8749 & 0.9341   & 0.8697     & 0.9007  & 0.4075  & 2.0567  & 0.6085    & 0.3463  & \textbf{0.6915}  \\
Mobile-VLM~\cite{chu2023mobilevlm}              & DD-VQA                  & 0.7699 & 0.7844   & 0.9081     & 0.8417 & 0.5202  & 3.0743  & 0.6778    & 0.3879  & 0.6544   \\
Mobile-VLM              & DD-VQA+                   & 0.8802 & 0.9451   & 0.9055     & 0.9249  & 0.5308  & 3.1776  & 0.6840    & 0.3922  & 0.6599  \\
MGFFD-VLM               & DD-VQA+ & \textbf{0.9072} & \textbf{0.9619} &\textbf{0.9228} & \textbf{0.9420} & \textbf{0.5349} & \textbf{3.3008} & \textbf{0.6963} & \textbf{0.3988} & 0.6629\\
\hline
\end{tabular}
}
\end{table*}

\begin{table*}[htb]
\centering
\caption{The performance of multi-modal enhanced deepfake detection. We report the AUC (\%). For the specially annotated "video," it represents the video-level result, while "frame" represents the frame-level result.}
\scalebox{0.8}{
\begin{tabular}{cccccc}
\hline
\multirow{2}{*}{Method} & \multirow{2}{*}{Multi-modal Enhancement} & \multicolumn{2}{c}{Celeb-DF} & DFDC                      & WDF                        \\
\cline{3-6}
                        &                                          & Frame & Video                & Frame                     & Frame                      \\
\hline
Xception~\cite{chollet2017xception}                & \ding{56}                                & 61.80 & -                    & 64.05                     & 62.72                      \\
Xception + PCC~\cite{hua2023learning}  & \ding{56}                                & 54.87 & -                    & 62.73                     & -                      \\
Xception + hierarchical~\cite{yu2024uncertainty} & \ding{56}                                & 72.86 & -                    & 69.23                     & -                      \\
Xception                & BLIP-TI                                  & 64.30 & -                    & -                         & 64.53                      \\
Xception                & MGFFD-VLM                                & 74.78 & -                    & 77.03                     & 77.84                     \\
\hline
RECCE~\cite{cao2022end} & \ding{56}                                & 68.71 & -                    & 62.41                     & 64.31                      \\
RECCE                   & BLIP-TI                                  & 70.21 & -                    & -                         & 69.46                      \\
RECCE                   & MGFFD-VLM                                & 70.77 & -                    & 72.67                     & 74.38                    \\
\hline
SBI~\cite{shiohara2022detecting} & \ding{56}                        & 86.12 & 93.18               & 74.44                     & 70.27                      \\
SBI                     & BLIP-TI                                  & -     & 93.98               & -                         & -                          \\
SBI                     & MGFFD-VLM                                & 87.80 & 93.53               & 81.77                     & 81.05                     \\
\hline
\end{tabular}}
\label{table:combined}
\end{table*}

\subsection{Forgery-Aware Training Strategy}
Training a VLM to detect subtle forgeries and explain them requires careful strategy. We employ a three-stage \textbf{forgery-aware training} scheme along with multiple loss functions to progressively tune the model:

\textbf{Stage 1: Multi-Granular Visual Learning.}
In the first stage, we focus on training the vision-side components for forgery detection. We freeze the LLM and train the segmentation decoder and the binary classification head using supervised signals from the images. We apply a binary cross-entropy loss $L_b$ on the image-level real/fake prediction $\hat{y}_i$ (using the ground truth label $y_i$), and a pixel-wise cross-entropy loss $L_s$ on the segmentation map $\hat{s}_i$ (using the ground-truth mask $s_i$ of manipulated regions, when available). For the blended images we generated, the ground-truth mask is the region that was replaced. Through $L_b$ and $L_s$, the model learns both coarse (image-level) and fine (pixel-level) features that distinguish fake content.

\textbf{Stage 2: Forgery Adaptation with Prompts.}
In the second stage, we unfreeze the image projector and the prompt-related parameters (the learnable prompt embeddings and the projector for the segmentation features). The LLM remains frozen. We now introduce the multi-granularity prompts and train the model to align these prompts with the textual answers. At this stage, we include all types of Q\&A from DD-VQA+ (common forgery questions, region/type, quality, etc.), so the prompt generation modules learn to assist in answering them correctly.

We also introduce a \emph{fine-grained contrastive loss} $L_{f}$ to reinforce the separation between real and fake in the feature space. As illustrated in Figure~\ref{fig:model-loss}, we treat a partially fake (blended) image as a hard positive example that should be closer to fully fake images than to real ones in the representation space. We form triplets consisting of a real image $P$, a fake image $N$, and a blended image $H$ (which contains some fake content). Let $sim(a,b)$ denote the cosine similarity between features of images $a$ and $b$. We impose:
\[ L_{f} = -\log \frac{\exp(sim(N, H))}{\exp(sim(N, H)) + \exp(sim(P, H))}, \]
along with standard terms to pull $P$ closer to $P$ and $N$ closer to $N$ (while pushing $P$ away from $N$). This loss encourages $H$ (the hard positive) to be more similar to $N$ (fake) than it is to $P$ (real), forcing the model to pick up even faint forgery cues in $H$. By the end of Stage 2, the visual encoder and prompt generators are well-aligned with forgery-related concepts, though the LLM itself has not yet been fine-tuned on these tasks.

\textbf{Stage 3: Forgery Instruction Tuning.}
In the final stage, we fine-tune the entire VLM (the LLM is now trainable, augmented by our LoRA modules) on the QA tasks. We feed in the full prompts + image + question and train the model to generate the correct answer text. We use the standard autoregressive language modeling loss $L_{text}$ on the answer sequences. Additionally, we introduce a \emph{text calibration loss} $L_{tcs}$ to ensure the model firmly distinguishes real vs fake in its textual output. Concretely, whenever the answer contains a phrase like It is a real (fake) face, we apply a binary cross-entropy on the logits of the word real vs fake at that position, using the ground truth label. This penalizes any ambiguity in the model's choice of the authenticity word. By explicitly training the model on this classification word, we prevent it from hedging (e.g., assigning high probability to both real and fake) and thus obtain clearer decisions in the generated explanations.

Through these three stages, our model first learns to extract and localize forgery features, then learns to represent and prompt them for the language model, and finally learns to articulate the detection and explanation in natural language. The total loss $L = L_{text} + \lambda_1 L_b + \lambda_2 L_s + \lambda_3 L_f + \lambda_4 L_{tcs}$ (with $\lambda$ coefficients to balance terms) is optimized over the entire training process (with each stage focusing on different subsets as described).

\begin{table*}
\centering
\caption{We present ablation experimental results for Multi-Granularity Prompt.The best results are shown in bold numbers. \textbf{Note:} FR+FT denotes Forgery Regions and Forgery Type; \textit{Loc Prompt} represents Location Prompt; \textit{Prob Prompt} stands for Probability Prompt. \textit{ADH LoRA} stands for Attribute-Driven Hybrid LoRA Strategy.}
\label{table:ablation2}
 \scalebox{0.75}{
\begin{tabular}{ccccccccccccc}
\hline
\multicolumn{2}{c}{DD-VQ+}          & \multicolumn{2}{c}{Prompt Compents}   & \multicolumn{4}{c}{Deepfake Detection}                                     & \multicolumn{5}{c}{Answer  Generation}                                                        \\
\hline
FR+FT &ADH LoRA & Loc  Prompt & Prob Prompt & Acc$\uparrow$   & Recall$\uparrow$ & Precision$\uparrow$ & F1$\uparrow$    & BLUE-4$\uparrow$ & CIDEr$\uparrow$ & ROUGE\_L$\uparrow$ & METEOR$\uparrow$ & SPICE$\uparrow$  \\
\hline
    &            &                  &                    & 0.8745          & 0.9260           & 0.9143     & 0.9201          & 0.5214           & 3.1439          & 0.6789             & 0.3891           & 0.6578           \\
\checkmark        &       &                  &                    & 0.8802          & 0.9451           & 0.9055              & 0.9249          & 0.5308           & 3.1776          & 0.6840             & 0.3922           & 0.6599           \\
\checkmark       &        & \checkmark                &                    & 0.8954          & 0.9427           & \textbf{0.9251}     & 0.9343          & 0.5411           & 3.2872 & \textbf{0.6989}    & 0.3959           & 0.6547           \\
\checkmark      &         & \checkmark                & \checkmark                  & 0.9030 & 0.9636  & 0.9168              & 0.9396 & \textbf{0.5482}  & 3.2380          & 0.6937             & 0.3973  & 0.6586  \\
\checkmark & \checkmark & \checkmark &\checkmark & \textbf{0.9072} & \textbf{0.9619} &0.9228 & \textbf{0.9420} & 0.5349 & \textbf{3.3008} & 0.6963 & \textbf{0.3988} & \textbf{0.6629}  \\
\hline
\end{tabular}
}
\end{table*}

\section{Experiments}

\subsection{Experimental Setup}

\textbf{Training Dataset.}
Our image training data mainly comes from FaceForensics++ (FF++)~\cite{rossler2019faceforensics++}. The text training data mainly includes the Deepfake Detection VQA dataset (DD-VQA~\cite{zhang2024common}), as well as three additional attributes generated by ourselves.


\textbf{Evaluation Dataset.}
Following the settings of Zhang \textit{et al.}~\cite{zhang2024common}, we focus on two aspects of the text responses on DD-VQA: the classification performance of the text answers and the generation performance of the text generation.
For text classification indicators, we evaluate the Accuracy (Acc), Recall, Precision, and F1. For text answering indicators, following the settings in~\cite{zhang2024common}, we evaluate metrics such as BLUE4~\cite{papineni2002bleu}, CIDEr~\cite{vedantam2015cider}, ROUGEL~\cite{lin2004rouge}, METEOR~\cite{banerjee2005meteor}, and SPICE~\cite{anderson2016spice}.
For evaluating our method enhanced deepfake detection methods, we choose several challenging face forgery datasets, namely Celeb-DF~\cite{li2020celeb} ,  Deepfake Detection Challenge (DFDC)~\cite{dolhansky2020deepfake}, and Wilddeepfake (WDF)~\cite{zi2020wilddeepfake}. We report the area under the receiver
operating characteristic curve (AUC).

\textbf{Implementation Details.}
We extract frames from each video and process face alignment. We resize all face images into $336\times 336$.
Our proposed method is based on the MobileVLM~\cite{chu2023mobilevlm}.
 We use Adam optimizer to train the framework with betas of $0.9$ and $0.995$.
 We set the batch size rate to $64$ for the first and second stages, and $48$ for the third stage.
 We set the learning rate to $4e-5$ for the first and second stages, and $1e-6$ for the third stage.
 We set the rank and alpha of LoRA to $64$ and $16$ respectively, and the LoRA dropout rate to $0.05$.

\begin{figure}
  \centering
   \includegraphics[width=0.6\linewidth]{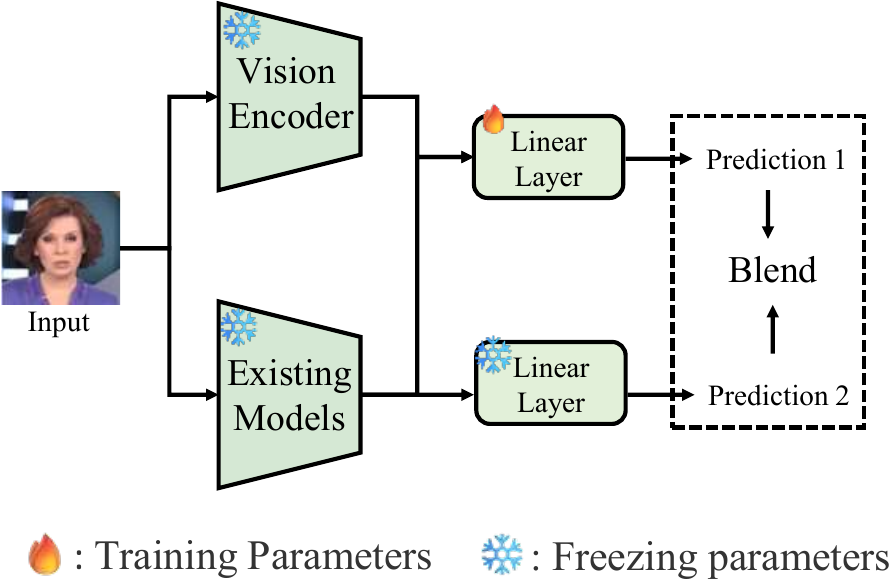}
   \caption{MGFFD-VLM Enhanced Deepfake Detection.
We combine the vision features of the large model with the existing features}
   \label{fig:model-enhabce}
\end{figure}
\subsection{Results on the DD-VQA dataset}

\begin{figure}
  \centering
   \includegraphics[width=1\linewidth]{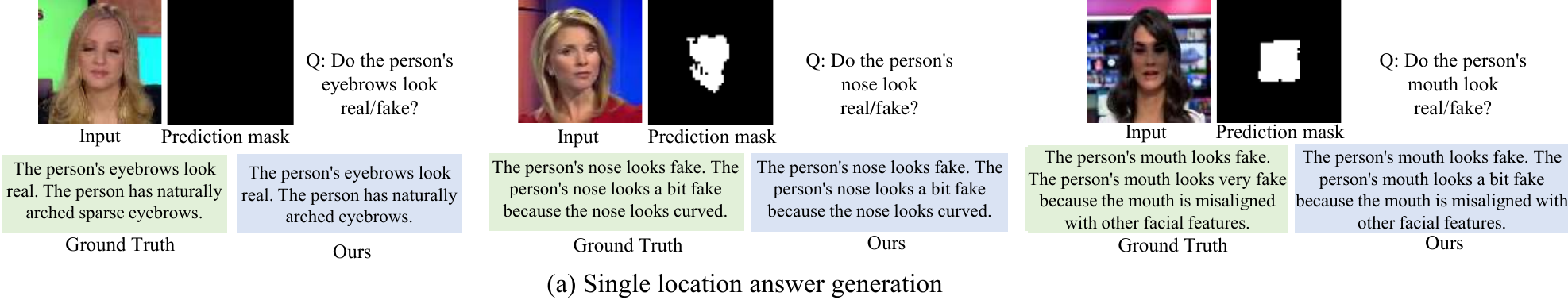}
   \includegraphics[width=1\linewidth]{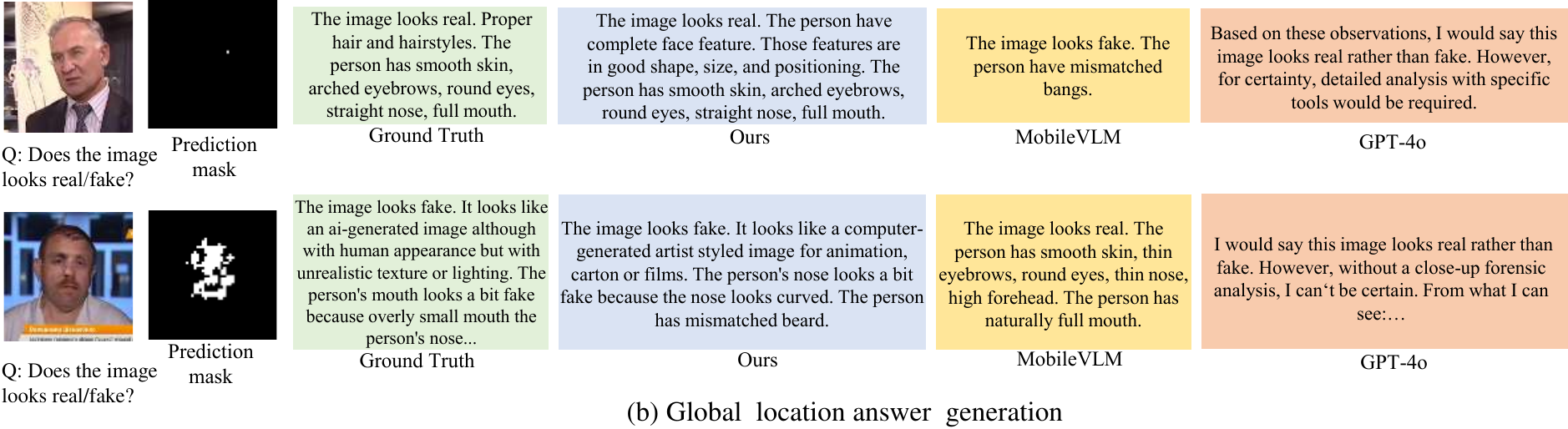}
   \caption{Results of Qualitative Examples. }
   \label{fig:Qualitative}
\end{figure}

We fine-tune MobileVLM and our proposed MGFFD-VLM on DD-VQA+ datasete and provide results for both deepfake detection and answer generation.
Our model can according to the template answer \textit{"it is real/fake"}  and provide the corresponding reasons. We conduct corresponding comparison experiments and ablation experiments to illustrate the role of our proposed DD-VQA+ and multi-granularity prompt learning. As shown in Table~\ref{table:compare1},
Row\#1 and Row\#2 are the results of fine-tuning BLIP using the DD-VQA dataset, and on this basis, the contrastive loss between images and texts is added.
Row\#3 and Row\#4 are both based on Mobile-VLM. The former uses DD-VQA while the latter uses DD-VQA+. It can be found that both have good performance in answer generation. However, due to the insufficient richness of DD-VQA+, in the deepfake detection task, using DD-VQA+ increases the Acc by 10\% compared to DDVQA.
Row\#5 is based on Mobile-VLM. By adding the Probability prompt and Location prompt, it can be seen that the Acc of this method increases by more than 2\%, and at the same time, the language ability also improves significantly.
\begin{table}
\centering
\caption{We present ablation experimental results for loss. The best results are shown in bold numbers.}
\label{table:ablation3}
\scalebox{0.8}{
\begin{tabular}{cccccc}
\hline
\multirow{2}{*}{$L_f$}  &  \multirow{2}{*}{$L_{tcs}$} & \multicolumn{4}{c}{Deepfake Detection}                \\
\cline{3-6}
                                           &                           & Acc$\uparrow$    & Recall$\uparrow$ & Precision$\uparrow$ & F1$\uparrow$      \\
\hline
                                   &                             & 88.32 & 94.73  & 90.74     & 92.69  \\
\checkmark                                                 &             & 89.22 & 95.11  & 91.48     & 93.26  \\
\checkmark           & \checkmark & \textbf{90.64}         & \textbf{95.92}            & \textbf{92.41}               & \textbf{94.13} \\
\hline
\end{tabular}
}
\end{table}



\subsection{Results on Deepfake Detection Models}
As shown in Figure~\ref{fig:model-enhabce}, we combine the vision features of the large model with the existing features to assess whether this fusion can enhance the effectiveness of existing models in forgery detection. We blend the output prediction from the small model (Prediction 1) with the fused result (Prediction 2) to obtain the final prediction.
As shown in Table~\ref{table:combined}, we conduct experiments on three classic methods: Xception~\cite{chollet2017xception}, RECCE~\cite{cao2022end} and SBI~\cite{shiohara2022detecting}.
For Xception, our method demonstrates significant improvement. Specifically, the detection auc improved by 12.98\% on Celeb-DF, 12.98\% on DFDC, and 15.12\% on WDF.
For RECCE, our method demonstrates significant improvement. Specifically, the detection auc improved by 2.06\% on Celeb-DF, 10.26\% on DFDC, and 10.07\% on WDF.
For SBI, the original paper primarily presents results at the video level. To ensure comparability, we first provide video-level detection results for Celeb-DF. Our method achieves a significantly higher improvement of 0.35\%. Additionally, we report frame-level results for SBI on Celeb-DF, WDF, and DFDC. Our reproduced results align closely with the official values provided in the original study. Notably, our method shows considerable improvement on Celeb-DF and WDF, with a minor performance decline on DFDC.
Overall, these results demonstrate that our proposed method effectively enhances the detection performance of existing deepfake detection models across multiple datasets.

\subsection{Results of Ablation Experiments}
To explore how the designed key modules affect model performance, we conducted ablation experiments.

\textbf{Effect of Forgery Regions and Forgery Types.} As shown in Table~\ref{table:ablation2},
Row\#1 represents the baseline trained only with the DD-VQA dataset. Row\#2 supplements Forgery Regions and Forgery Types on the basis of the baseline.
Incorporating the online-generated forgery localization task significantly enhances both the detection and response capabilities of the VLM. This improvement is attributed to the task's focus on fine-grained forgery areas, which helps the VLM better understand forgery-related features.

\textbf{Effect of Multi-Granularity Prompt.}
As presented in Table~\ref{table:ablation2},
Row\#3 and Row\#4 are the experimental results of adding Location Prompt and Probability Prompt in sequence.
Adding the Location Prompt increases accuracy by 1.52\% over Row\#2, along with comprehensive improvements in language metrics. This indicates that incorporating forgery-related prompts enhances the model's response capabilities. With the addition of the Probability Prompt, accuracy increases by 0.76\%, though improvements in answer generation are less noticeable since the Probability Prompt is primarily suited to classification tasks. This prompt has less impact when the model is asked to identify unnatural aspects in specific regions. Overall, adding both Location and Probability Prompts achieves the best performance.

\textbf{Effect of Attribute-Driven Hybrid LoRA Strategy.}
As presented in Table~\ref{table:ablation2},
Row\#5 are the experimental results of adding Attribute-Driven Hybrid LoRA Strategy. Our LoRA integration led to significant gains in numerical metrics (e.g., deepfake detection accuracy).
Simultaneously, the performance of most text - related tasks also shows an enhancement. This improvement can be attributed to the selection of corresponding experts for photos of different qualities, which in turn enhances the detection capabilities. Regarding the ablation experiments between the image feature and the quality feature, please refer to the supplementary file for more details.

\textbf{Effect of Fine-Grained Loss and Text Cross-entropy Loss.}
As shown in Table~\ref{table:ablation3}, Row\#1 represents the MGFFD-VLM model without fine-grained loss and text cross-entropy loss during entire training phase. Row\#2 builds upon the model of  Row\#1 by adding fine-grained loss, and Row\#3 further enhances the model of Row\#2 with the addition of text cross-entropy loss.
when the fine-grained loss is added, the Acc of the model's deepfake detection rises from 88.32\% to 89.22\%. This shows that our fine-grained loss can help the model learn fine-grained features that are difficult to mine. At the same time, after adding the Text cross-entropy loss, the Acc rises by 1.42\%, indicating that multiple auxiliary losses can simultaneously promote forgery detection.



Qualitative results are provided in Figure~\ref{fig:Qualitative}. In these examples, our model accurately identifies fake faces and provides clear explanations (e.g., pointing out unnatural skin texture or odd facial blending in the fake images, or noting a completely natural appearance for a real image). In contrast, a generic large model (GPT-4) might only give a vague or guess-based answer, and the baseline MobileVLM without our enhancements often fails to pinpoint the issues. Our MGFFD-VLM can both detect the forgery and articulate the specific evidence, demonstrating the effectiveness of our approach.


\section{Conclusion}
We presented \textbf{MGFFD-VLM}, a vision-language framework for face forgery detection that produces both decisions and human-understandable explanations. By extending the training data (DD-VQA+) with richer forgery-related questions and by incorporating multi-granularity prompts, quality-aware LoRA experts, and a tailored training strategy, our model achieves state-of-the-art accuracy on deepfake detection while providing clear rationales. The results demonstrate the promise of using large VLMs for interpretable deepfake detection. In future work, we plan to explore further improvements in generalization to unseen manipulation methods.

\bibliography{aaai2026}


\end{document}